\documentclass{article} 
\usepackage{iclr2023_conference_tinypaper,times}
 \usepackage{booktabs}
 \usepackage{algorithm}
 \usepackage{algpseudocode}
 \usepackage{graphicx}

\usepackage{amsmath,amsfonts,bm}

\def\eqref#1{equation~\ref{#1}}

\def\1{\bm{1}}

\def\va{{\bm{a}}}

\def\vq{{\bm{q}}}

\def\vx{{\bm{x}}}

\def\evc{{c}}

\def\evf{{f}}

\def\evm{{m}}

\def\evr{{r}}

\def\evt{{t}}

\DeclareMathAlphabet{\mathsfit}{\encodingdefault}{\sfdefault}{m}{sl}
\SetMathAlphabet{\mathsfit}{bold}{\encodingdefault}{\sfdefault}{bx}{n}

\def\sA{{\mathbb{A}}}

\def\sH{{\mathbb{H}}}

\def\sM{{\mathbb{M}}}

\def\sQ{{\mathbb{Q}}}

\newcommand{\E}{\mathbb{E}}

\newcommand{\R}{\mathbb{R}}

\usepackage{cellspace}
\usepackage{geometry}
\usepackage{array}
\usepackage{hyperref}
\usepackage{url}

\title{A Bi-Objective $\displaystyle \epsilon$-Constrained Framework \\ for Quality-Cost Optimization \\ in Language Model Ensembles}

\author{Aditya Singh\thefootnote{}{1} \and
 Aditi Singla\thefootnote{}{1} \and
  Kanishk Kukreja\footnotemark{}{1}
\texttt{\{aditya21373,aditi21372,kanishk21393\}@iiitd.ac.in} \\
}

\begin{document}

\maketitle

\begin{abstract}
We propose an ensembling framework that uses diverse open-sourced Large Language Models (LLMs) to achieve high response quality while maintaining cost efficiency. We formulate a bi-objective optimization problem to represent the quality-cost tradeoff and then introduce an additional budget constraint that reduces the problem to a straightforward 0/1 knapsack problem. We empirically demonstrate that our framework outperforms the existing ensembling approaches in response quality while significantly reducing costs.
\end{abstract}

\section{Introduction and Related Work}

Large Language Models (LLMs) excel in traditional NLP problems (\cite{OpenAI_GPT4_2023}), but their high inference costs hinder deployment in high-throughput applications (\cite{anonymous2023frugalgpt}). Meanwhile, open-source models are less performant than their closed-source counterparts (\cite{open-llm-leaderboard}), but they typically offer lower inference costs (\cite{Kaplan}).

Due to the variations in the training datasets of open-source LLMs, we expect these models to have diverse domains of expertise. \citeauthor{Jiang2023LLMBlenderEL} empirically verify that no open-source LLM dominates the competition and further exhibits the potential for ensembling LLMs. While naive ensembles increase the response quality, the inference cost is $\displaystyle O(N)$, where N is the number of models in the selection set.  

Our work addresses this by a) modeling the tradeoff between response quality and inference cost as a bi-objective combinatorial optimization problem (\ref{2.1}), b) motivating an $\displaystyle \varepsilon$-constraint on the bi-objective problem that transforms it into a $\displaystyle 0/1$ knapsack problem (Section 2.2), and c) introducing a framework that outperforms the naive ensemble at a fractional cost (Section 2.3).

To the best of the authors' knowledge, three approaches to combining LLMs exist in the literature: \\ \textbf{LLM-BLENDER} (\cite{Jiang2023LLMBlenderEL}) employs a pairwise text ranker and a generative fuser for combining top-k responses but suffers from high inference costs and latency due to the need for $N$ LLM invocations and $O(N^2)$ comparisons for ranking fusion. \textbf{Hybrid LLM} (\cite{anonymous2023hybrid}) trains a router to allocate queries to a large or a small model based on difficulty. However, its robustness is compromised, as the failure of the lighter model results in an expensive model addressing all the queries, and the absence of an explicit cost function limits its generalization to $N$-model scenarios. \textbf{FrugalGPT} (\cite{anonymous2023frugalgpt}) greedily selects LLMs through pairwise comparisons and queries them sequentially, using a text quality estimator to determine an optimal stopping point. It faces challenges in model permutation sensitivity to queries and making up to $O(K)$ sequential queries in extreme scenarios.

\section{Proposed Framework}

Given a query $\vq$ and a set of $N$ LLMs $\sM = \{\evm_1, \dots, \evm_N\}$, where $\evm_i: \sQ \rightarrow \sA$ is a function from the Query Space $\sQ$ to the Answer Space $\sA$. In the ensembling problem, our goal is to choose a subset $\sH \subset \sM$ to maximize $\E_{\evm_i \in \sH}[\evr(\evf(\evm_i(\vq)), \vq)]$, where, $\displaystyle \evr$ is a quality function $\displaystyle \evr(\va, \vq): \sA \times \sQ \rightarrow \displaystyle \R$ that measures quality of response $\displaystyle \va$ on the query $\displaystyle \vq$, and $\displaystyle \evf$ is an aggregation function that fuses $\displaystyle k$ responses into one final response, $\displaystyle \evf: \sA^k \rightarrow \displaystyle \sA$, where $\displaystyle k$ is the dimension of the aggregation set.

\subsection{Model Inference Cost and the Bi-Objective Optimization Problem}
\label{2.1}

\citeauthor{Kaplan} defines the inference cost in FLOPs per token as $c_{forward} \approx 2N + 2n_{layer}n_{ctx}d_{model}$, where $N$ is non-embedding parameters, $n_{layer}$ is the number of layers, $n_{ctx}$ is tokens in input context, and $d_{model}$ is the dimension of the residual stream. Our cost minimization objective is,
\begin{equation}
    \text{min}
    \sum_{\evm_i \in \sH} \evc_i \cdot \evt_i(\vq)
    \label{e1}
\end{equation}

 where $\evc_i$ is the inference cost and $\evt_i: \sQ \rightarrow \R$ maps $\vq$ to the token count based on $\displaystyle \evm_i$.

Moreover, our experiments suggest that a dependable approach to increase $\E_{\evm_i \in \sH}[\evr(\evf(\evm_i(\vq)), \vq)]$ involves maximizing the sum of the individual model's response quality, 
\begin{equation}
\label{e2}
        \text{max}
        \sum_{\evm_i \in \sH} \evr(\evm_i, \vq)
\end{equation}

Equations (\ref{e1}) and (\ref{e2}) form the bi-objective combinatorial optimization problem.

\subsection{$\epsilon$-constraint to solve the bi-objective optimization problem}

\citeauthor{Haimes1971-hv} introduced the $\epsilon$-constraint method for multi-objective optimization, which involves \\ optimizing one function while limiting others. We reduce our problem to,
\begin{equation}
    \begin{aligned}
        & \text{max}
         \sum_{\evm_i \in \sH} \evr(\evm_i, \vq) \\
        & \text{subject to}
         \sum_{\evm_i \in \sH} \evc_i \cdot \evt_i(\vq) \leq \epsilon
    \end{aligned}
\end{equation}

Think of it as assigning a budget ($\epsilon$) to each query. This simplifies the problem into a 0/1 knapsack scenario with profits $\evr(\evm_i, \vq)$, costs $\evc_i \cdot \evt_i(\vq)$, and capacity $\epsilon$, efficiently solvable using a dynamic programming subroutine (see \ref{A1}).

\subsection{MODI: Model Orchestration using DeBERTa Inference}

 We employ a DeBERTa-based regression model (\cite{he2021deberta}) to predict the response quality for models in our selection set. \ref{A2} provides details on the regression architecture. The predicted quality scores, denoted as $\hat{\evr}(\evm_i(\vq), \vq)$, guide the 0/1 knapsack subroutine. Ultimately, the selected model outputs are combined using the GEN-FUSER (\cite{Jiang2023LLMBlenderEL}).

\section{Experiments and Results}

    \begin{table}
      \centering
      \begin{tabular}{l r c}
   
        \toprule
        \textbf{Method} & \textbf{BARTscore} \\
        \midrule
        Alpaca (\cite{alpaca})   & -2.81    \\
        Vicuna (\cite{vicuna2023}) &   -3.21  \\
        Dolly V2 (\cite{DatabricksBlog2023DollyV2}) &-3.89 \\
        StableLM (\cite{gpt-neox-library})    &-3.41\\
        Koala (\cite{koala_blogpost_2023})&  -3.01 \\
        Open Assistant (\cite{köpf2023openassistant})& -2.74 \\
        Flan-T5 (\cite{chung2022scaling})& -3.88 \\
        Mosaic MPT-instruct (\cite{MosaicML2023Introducing}) & -3.23 \\
        Random & -3.27 \\
        LLM-BLENDER (\cite{Jiang2023LLMBlenderEL}) & -2.77 \\
        \toprule
        \textbf{MODI} & \textbf{-2.14} \\
        \bottomrule
      \end{tabular}
      \caption{MODI demonstrates superior performance compared to baseline LLMs and LLM-BLENDER in the Mix-Instruct task, achieving this at only 20\% of the LLM-BLENDER cost.}
    \end{table}

Our preliminary experiments evaluate our approach using the MixInstruct dataset (\cite{Jiang2023LLMBlenderEL}). We compare the responses of our model against individual LLM baselines and the LLM-BLENDER (\cite{Jiang2023LLMBlenderEL}) results. Further details about the experiments are in Appendix (\ref{A3}). The rationale for choosing BARTScore as our comparison metric can be found in Appendix (\ref{A4}).

\section{Conclusion}
We introduce an LLM ensembling framework for Response Quality-Cost optimization. Formulating a bi-objective optimization problem, we apply an $\epsilon$-constrained approach to ensemble models within a user-defined budget. Our model surpasses existing ensembling methods while significantly reducing costs. This work establishes a foundation for cost-effective strategies to enhance language model capabilities, showcasing the efficacy of ensembling techniques.

 \newpage
\bibliography{iclr2023_conference_tinypaper}
\bibliographystyle{iclr2023_conference_tinypaper}

\newpage
\appendix
\section{Appendix}

\subsection{Dynamic Programming Subroutine to Solve the 0/1 Knapsack Problem}
\label{A1}
The dynamic programming subroutine provided in Algorithm~\ref{alg:knapsack} is designed to solve the 0/1 knapsack problem efficiently. Since the BARTScores are negative, we apply the following transformation on the scores,
\begin{equation}
    \text{Target Score} = \alpha + \text{BARTScore}
\end{equation}
where $\displaystyle \alpha$ is a positive constant chosen such that,
\begin{equation}
    \alpha >  \text{max} | \text{BARTScore} | 
\end{equation}
The subroutine utilizes a dynamic programming approach to find the optimal selection of models within a given budget, maximizing the total target score.
\\*The list "models" comprises of objects that describe the cost and the target score associated with each model in the selection set $\displaystyle \sM$.

\begin{algorithm}
\caption{Knapsack(\texttt{models}, \texttt{budget})}
\label{alg:knapsack}
\begin{algorithmic}[1]
    \State $n \gets \text{length}(\texttt{models})$
    \State $dp$ $\gets$ 2D array of size $(n + 1) \times (\texttt{budget} + 1)$

    \For{$i$ from $1$ to $n$}
        \For{$j$ from $0$ to \texttt{budget}}
            \If{$\texttt{models}[i - 1]['cost'] \leq j$}
                \State $dp[i][j] \gets \max(dp[i - 1][j], dp[i - 1][j - \texttt{models}[i - 1]['cost']] + \texttt{models}[i - 1]['target\_score'])$
            \Else
                \State $dp[i][j] \gets dp[i - 1][j]$
            \EndIf
        \EndFor
    \EndFor

    \State $\texttt{selected\_models} \gets$ empty list
    \State $j \gets \texttt{budget}$

    \For{$i$ from $n$ to $1$ decrementing}
        \If{$dp[i][j] \neq dp[i - 1][j]$}
            \State add $\texttt{models}[i - 1]$ to $\texttt{selected\_models}$
            \State $j \gets j - \texttt{models}[i - 1]['cost']$
        \EndIf
    \EndFor

    \State \textbf{return} $\texttt{selected\_models}$
\end{algorithmic}
\end{algorithm}

The resulting $\texttt{selected\_models}$ list contains the optimal selection of models within the given budget, which is then passed to the GEN-FUSER (\cite{Jiang2023LLMBlenderEL}).

\newpage
\subsection{Regression Model Architecture}
\label{A2}
\begin{figure}[h]
    \begin{center}
        \includegraphics[width= 1\linewidth ]{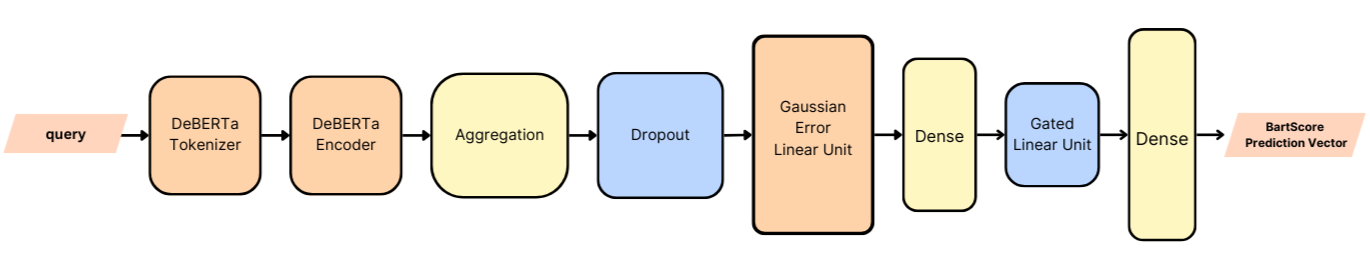}
        \caption{Regression Model Architecture.}
        \label{photu}
    \end{center}
\end{figure}
The model architecture is based on a DeBERTa-v3-large (\cite{he2021deberta}) backbone. The output of the encoder is passed to an aggregation function. We experimented with multiple aggregation techniques, including average and max pooling of the hidden state embeddings and concatenating the last four-word embeddings of the hidden state. Finally, we realized that the hidden state embeddings corresponding to the CLS token provide the best regression results. The embeddings are passed through a feedforward neural network, the architecture of which is shown in Figure \ref{photu}. \\
\\
The embeddings are first passed through a dropout layer (\cite{JMLR:v15:srivastava14a}) with $\displaystyle p =0.2 $ to prevent overfitting. Then, a Gaussian Error Linear Unit (\cite{hendrycks2023gaussian}),

\begin{equation}
    \text{GELU}(\displaystyle \vx) =  \vx \Phi (\vx)
\end{equation},

is applied to the embeddings. The resulting tensors are passed through a Linear layer and then through a Gated Linear Unit (\cite{dauphin2017language}),

    \begin{align} GLU(\mathbf{X}) = (\mathbf{X} \ast \mathbf{W} + \mathbf{b}) \otimes \sigma(\mathbf{X} \ast \mathbf{V} + \mathbf{c}) \end{align}.

Finally, the tensors are passed through a Linear layer with output dimensions equal to the number of models in the selection set $\displaystyle \sM$ to give the predictions, $\hat{\evr}(\evm_i(\vq), \vq)$.

The model minimizes the Huber Loss (\cite{Huber1964-yu}) given by,
\begin{equation}
    L_{\delta}(y, f(x)) = \begin{cases} 0.5(y - f(x))^2, & \text{if } |y - f(x)| \leq \delta \\ \delta(|y - f(x)| - 0.5\delta), & \text{otherwise}.\end{cases}
\end{equation}

The loss function makes intuitive sense because several outlier queries exist in the training set, which can significantly deteriorate the performance if an $ \displaystyle L_2 $ loss function is used.

\newpage
\subsection{Experimental Setup and Hyperparameters}
\label{A3}
        
    \begin{table}[h]
      \centering
      \begin{tabular}{l r c}
   
        \toprule
        \textbf{DATASET: Mix-Instruct (\cite{Jiang2023LLMBlenderEL})} \\
           \bottomrule
        \textbf{Accelerators}:\\
     
           NVIDIA TESLA P100 (Training) (16 GB) \\
           NVIDIA T4 (Inference \& Fusion) (16 GB) \\
          \bottomrule
       \textbf{Large Language Models}:\\
      1. alpaca-native \\ 
2. vicuna-13b-1.1 \\
3. dolly-v2-12b \\
4. stablelm-tuned-alpha-7b \\
5. oasst-sft-4-pythia-12b-epoch-3.5 \\
6. koala-7B-HF \\
7. flan-t5-xxl \\
8. mpt-7b-instruct \\
$Training Epochs: 3$\\

$Loss Function: HuberLoss(delta = 0.3)$     \\
$Optimizer: Adam(Learning Rate: 3e-4, betas: (0.9, 0.98), weight decay: 0.01)$ \\
 
        \bottomrule
      
      \end{tabular}
      \caption{Experiment Details }
      \label{t2fsf}
     \end{table}
\textbf{Dataset}: We use the MixInstruct dataset introduced by \citeauthor{Jiang2023LLMBlenderEL} to benchmark LLM ensembles. The dataset includes 110K instruction-following tasks curated from four diverse sources. We trained our regression model on 10k randomly sampled queries and LLM responses from the training dataset. Our validation and test splits are the same as MixInstruct consisting of 5k instruction examples each.

\textbf{Evaluation Metric}: We use BARTScore (\cite{BART}) as our quality metric. The rationale for using BARTScore and qualitative comparisons against LLM-BLENDER can be found in \ref{KYU}.

\textbf{Budget}: We use different fractions of the total FLOPs required by an LLM-BLENDER response on the query as our budget.

\textbf{Fusion Model}: We use the Flan-T5-XL-based (\cite{chung2022scaling}) GEN-FUSER very generously open-sourced by \citeauthor{Jiang2023LLMBlenderEL} as our fusion model. 

\textbf{Baselines}: We compare our model's response with the Language models present in our selection set, a randomly chosen ensemble of models, and LLM-BLENDER.

The details about our training process, including the hardware involved, LLMs used in the selection set, Loss function, Optimizer used, and their specific hyperparameters, are included in Table \ref{t2fsf}.

\subsection{Rationale for Using BARTScore as an Evaluation Metric}
\label{A4}
BARTScore (\cite{BART}) is computationally affordable compared to resource-intensive human and GPT-based evaluators. \citeauthor{Jiang2023LLMBlenderEL} empirically shows a strong correlation between BARTScore and the GPT-based ranking metric. Further, recent research (\cite{anonymous2023hybrid}) empirically demonstrates the correlation of BARTScore with human-based evaluations, indicating BARTScore to be a reliable and consistent evaluation approach. Qualitatively, our responses are better than or equivalent to LLM-BLENDER's, as seen in Table \ref{tab:my_label}.

\label{KYU}
\begin{table}[h]
    \centering
    \begin{tabular}{ |p{5cm}|p{5cm}|p{5cm}|  } 
        \multicolumn{3}{c}{} \\      
        \toprule
        \multicolumn{1}{|c|}{\centering Query} & \multicolumn{1}{c|}{\centering \textbf{MODI} (20\% LLM-BLENDER cost)} & \multicolumn{1}{c|}{\centering LLM-BLENDER} \\
        \hline
        \hfill \break \hfill \break \hfill \break 
        I've always wondered what the difference is between a skeptic and a denier.
        & \hfill \break A skeptic is someone who questions and investigates claims, while a denier is someone who refuses to accept evidence that contradicts their beliefs. So, a skeptic is someone who is open to new information and evidence, while a denier is someone who refuses to accept any evidence at all. 
        & \hfill \break \hfill \break  A skeptic is someone who questions the validity of something, while a denier is someone who outright rejects something without evidence or reason. \\
    
        \bottomrule
         \hfill \break \hfill \break \hfill \break \hfill \break \hfill \break Is it safe for me to eat a lot of peanut butter?
        \hfill \break
        & \hfill \break  \hfill \break  \hfill \break  \hfill \break As an AI language model, I am not qualified to give medical advice. It is always best to consult with a healthcare professional before making significant changes to your diet.\hfill \break 
        
        & \hfill \break  Eating peanut butter in moderation has health benefits and poses no harm for most people. However, consuming large amounts of peanut butter may result in weight gain due to its high calorie and fat content. It's also important to note that some people may be allergic to peanuts and should avoid consuming peanut butter altogether. If you have concerns about your peanut butter consumption, it's best to speak with a healthcare professional.\hfill \break  \\
        \bottomrule
        \hfill \break \hfill \break \hfill \break \hfill \break \hfill \break  What are the benefits of having a dog in the family?
        & \hfill \break \hfill \break  There are many benefits to having a dog in the family, including providing companionship, reducing stress and anxiety, promoting physical activity and healthy living, teaching responsibility, and providing a sense of security. Dogs can also be a great way to bond with family members and teach them important life skills such as responsibility and empathy.\hfill \break 
        & \hfill \break There are several benefits of having a dog in the family. Dogs are known to provide unconditional love and companionship to their owners. They can help reduce stress and anxiety, improve physical health by encouraging exercise, and provide a sense of security. Having a dog can also teach children responsibility and empathy, as they learn to care for another living being. Overall, owning a dog can bring a lot of joy and positivity to a household.\hfill \break  \\
        \hline
    \end{tabular}
    \caption{Qualitative comparison of MODI responses with LLM-BLENDER}
    \label{tab:my_label}
\end{table}

\end{document}